\documentclass[letterpaper]{article} 
\usepackage[submission]{aaai2026}  
\usepackage{times}  
\usepackage{helvet}  
\usepackage{courier}  
\usepackage[hyphens]{url}  
\usepackage{graphicx} 
\usepackage{natbib}  
\usepackage{caption} 
\usepackage{algorithm}
\usepackage{algorithmic}
\usepackage{amsmath} 
\usepackage{xcolor}
\usepackage{amsfonts}

\usepackage{booktabs}
\usepackage{multirow}
\usepackage{microtype}
\usepackage{pdfpages}
\usepackage{newfloat}
\usepackage{listings}
\DeclareCaptionStyle{ruled}{labelfont=normalfont,labelsep=colon,strut=off} 
\floatstyle{ruled}
\newfloat{listing}{tb}{lst}{}
\floatname{listing}{Listing}
\title{ViTE: Virtual Graph Trajectory Expert Router for Pedestrian Trajectory Prediction}

\author{
    Ruochen Li\textsuperscript{\rm 1}, 
    Zhanxing Zhu\textsuperscript{\rm 2}, 
    Tanqiu Qiao\textsuperscript{\rm 1}, 
    Hubert P. H. Shum\textsuperscript{\rm 1}\thanks{Corresponding author}
}

\affiliations{
    \textsuperscript{\rm 1}Department of Computer Science, Durham University, UK\\

    \textsuperscript{\rm 2}School of Electrical and Computer Science (ECS), University of Southampton, UK\\

    \{ruochen.li, hubert.shum\}@durham.ac.uk\\

}

\begin{document}

\maketitle

\begin{abstract}
Pedestrian trajectory prediction is critical for ensuring safety in autonomous driving, surveillance systems, and urban planning applications. While early approaches primarily focus on one-hop pairwise relationships, recent studies attempt to capture high-order interactions by stacking multiple Graph Neural Network (GNN) layers. However, these approaches face a fundamental trade-off: insufficient layers may lead to under-reaching problems that limit the model's receptive field, while excessive depth can result in prohibitive computational costs. We argue that an effective model should be capable of adaptively modeling both explicit one-hop interactions and implicit high-order dependencies, rather than relying solely on architectural depth. To this end, we propose ViTE (Virtual graph Trajectory Expert router), a novel framework for pedestrian trajectory prediction. ViTE consists of two key modules: a Virtual Graph that introduces dynamic virtual nodes to model long-range and high-order interactions without deep GNN stacks, and an Expert Router that adaptively selects interaction experts based on social context using a Mixture-of-Experts design. This combination enables flexible and scalable reasoning across varying interaction patterns. Experiments on three benchmarks (ETH/UCY, NBA, and SDD) demonstrate that our method consistently achieves state-of-the-art performance, validating both its effectiveness and practical efficiency.
\end{abstract}

\section{Introduction}
Human trajectory prediction aims to forecast future pedestrian paths based on observed motion histories. It plays a vital role in autonomous driving for tasks such as collision avoidance and emergency braking \cite{bai2015pomdpintro,luo2018porca,liu2021trajectorysurvey}, and is also essential in video surveillance for identifying suspicious activities \cite{luber2010videosur, shi2021sgcn}. This task is challenging due to the uncertainty of human behavior and the varying relevance of surrounding individuals: nearby agents may not interact, while distant ones can still be coordinated. This complexity requires models that can flexibly capture interactions across multiple scales.

\begin{figure}[t]\centering
  \includegraphics[width=\linewidth]{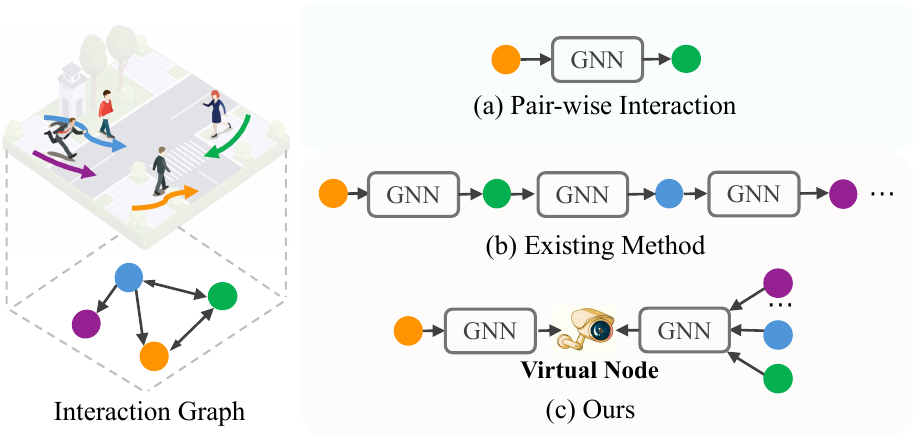}
  \caption{Comparison of interaction modeling strategies. \textbf{(a)} Traditional methods capture only one-hop interactions. \textbf{(b)} Existing methods stack multiple GNN layers to model high-order dependencies. \textbf{(c)} Our method introduces virtual nodes to capture high-order interactions efficiently.}
  \label{fig:teaser}
\end{figure}

Early approaches primarily focus on pairwise interactions, capturing local spatial dependencies between individuals. Representative methods like Social-LSTM \cite{Alexandre2016lstm,gupta2018socialgan} utilize social pooling to aggregate information from neighboring agents within a fixed window, while SS-LSTM \cite{xue2018ss-lstm} introduces occupancy maps to encode nearby spatial configurations. With the advancement of relational modeling, graph-based methods \cite{huang2019stgat,kosaraju2019socialbigat,Mohamed2020socialstgcnn,shi2021sgcn} have become prominent, representing pedestrians as nodes and their interactions as edges in a graph, enabling data-driven learning of pairwise relationships via Graph Neural Networks (GNNs) as shown in Figure \ref{fig:teaser} (a). To capture richer interaction dynamics, more recent efforts have focused on modeling high-order interactions—influences that are not directly observable between agent pairs but emerge through multi-hop connections within a crowd. Methods such as HighGraph \cite{kim2024higher} and PCHGCN \cite{chen2025pchgcn} address this by stacking multiple GNN layers (Figure \ref{fig:teaser} (b)), allowing information to propagate across longer distances and indirectly connected agents. These models have shown improvements in capturing group behavior patterns and non-local dependencies.

Despite their effectiveness, high-order interaction modeling via deep GNNs introduces two key challenges. The first is depth-related inefficiency. Stacking too few GNN layers results in under-reaching, where the receptive field is insufficient to cover relevant agents beyond local neighborhoods, leading to incomplete interaction modeling \cite{lu2024nodemixup}. Conversely, increasing depth can cause over-smoothing \cite{rusch2023surveyoversmoothing}, where repeated message passing dilutes node-specific features, making individual agent representations less discriminative. This trade-off poses a fundamental bottleneck for graph-based trajectory models that rely solely on depth to model relational complexity. Another major limitation is the lack of contextual adaptability. In real-world crowds, the influence of other pedestrians varies across scenes and situations: in some cases, immediate neighbors dominate decision-making, while in others, long-range or indirect interactions play a critical role. However, most existing methods apply a fixed, shared relational structure or aggregation scheme to all agents, ignoring this diversity. As a result, they fail to dynamically adjust the reasoning process according to individual agents’ social context and interaction scale.

To address these limitations, we propose ViTE (Virtual graph Trajectory Expert router), a novel framework for pedestrian trajectory prediction. It comprises two key components. First, the \textbf{Virtual Graph}, introduces dynamically assigned virtual nodes that act as relational mediators between pedestrians (Figure \ref{fig:teaser} (c)). These nodes serve as intermediate hubs for message exchange, enabling the model to capture long-range dependencies and high-order interactions without deep GNN stacks. This compact and learnable structure not only mitigates under-reaching but also improves efficiency and expressiveness in modeling long-range relations. Second, the \textbf{Expert Router} adopts a Mixture-of-Experts (MoE) design to enable context-aware interaction reasoning. We introduce multiple interaction experts, each specialized in a particular interaction scale (e.g., one-hop or high-order), are coordinated by a gating network that dynamically routes each agent’s representation to the most relevant experts. This adaptive mechanism allows ViTE to flexibly integrate diverse relational patterns based on social context. Together, these two modules form a cohesive and scalable system that overcomes depth limitations and enables fine-grained social reasoning for trajectory prediction. 
Our main contributions are:
\begin{itemize}
    \item We propose \textbf{ViTE}, a novel framework for pedestrian trajectory prediction that enables efficient and adaptive high-order interaction modeling.
    
    \item We design a \textbf{Virtual Graph} module that introduces dynamic virtual nodes as relational hubs to capture indirect social dependencies.
    
    \item We develop an \textbf{Expert Router} based on a Mixture-of-Experts mechanism to perform context-aware reasoning across multiple interaction scales.
\end{itemize}
\textbf{Code:} https://github.com/Carrotsniper/ViTE

\section{Related Work}
\subsection{Interaction Modeling for Trajectory Prediction}
Early work on pedestrian trajectory prediction focused on pairwise interactions to capture the influence of nearby agents. Social-LSTM \cite{Alexandre2016lstm, gupta2018socialgan} introduced a pooling mechanism to aggregate neighboring hidden states, while SS-LSTM \cite{xue2018ss-lstm} encoded spatial layouts via occupancy maps to enhance local interaction modeling. However, these methods struggle with complex multi-agent dynamics due to their limited relational structure. To address this, graph-based approaches \cite{bae2021disentangled,qiao20222ggcn,bae2023graphtern,qiao2024category} have been proposed, representing pedestrians as nodes and their interactions as edges in a graph. ST-GAT \cite{huang2019stgat} employs graph attention \cite{qiao2025geometricGNN,shao2025rethinkbraintumor,shao2025trace} to adaptively weigh neighboring influences, Social-STGCNN \cite{Mohamed2020socialstgcnn} combines spatial and temporal convolutions, Social-BiGAT \cite{kosaraju2019socialbigat} models bidirectional influence flows, and SGCN \cite{shi2021sgcn} applies sparse GCNs to capture spatial-temporal dependencies. These methods advance beyond pairwise designs by leveraging GNNs to jointly model spatial and temporal dependencies. To capture high-order interactions, which refer to indirect influences mediated through intermediate agents, recent works such as HighGraph \cite{kim2024higher} and PCHGCN \cite{chen2025pchgcn} stack multiple GNN layers to enable multi-hop message passing. While effective for modeling non-local dependencies, this strategy faces a trade-off: shallow networks suffer from under-reaching \cite{lu2024nodemixup,li2025unifiedTraj}, while deeper ones risk over-smoothing, degrading representation quality \cite{rusch2023surveyoversmoothing}. To this end, we introduce adaptive virtual nodes that serve as global relational hubs, effectively capturing high-order interactions by mediating indirect dependencies without relying on deep GNNs.

\subsection{Mixture of Expert}
Mixture of Experts (MoE) is a modular neural architecture that partitions the input space and routes each input to a subset of specialized experts, selected dynamically by a gating network \cite{MOE_survey,jiang2024interpretable,mu2025comprehensiveMOEreview}. Unlike traditional ensembles, MoE activates only a few experts per input, enabling high model capacity with reduced computational cost. This design has shown success in NLP \cite{adaptive_moe,outrageously_lnn}, vision \cite{wang2020deepmixoe,riquelme2021scaling}, and multi-modal learning \cite{mustafa2022multimodalmoe}. In graph learning, MoE has been applied to aggregate across neighborhoods \cite{abu2020nMOE}, correct bias \cite{hu2021graphMOEClass}, and enhance molecular prediction \cite{kim2023learningmoe}. Recent works explore top-$k$ input routing \cite{zhou2022mixtureofexperts} and multi-hop fusion \cite{wang2023graphMOE}, but often rely on fixed routing or task-specific designs, limiting adaptability. In this work, we propose a MoE-based expert router for trajectory prediction, which enables context-aware routing over one-hop and high-order graph interactions. By dynamically selecting the most relevant expert for each node, our model effectively adapts to diverse interactions in pedestrian crowds.

\section{Methodology}

\begin{figure*}[t]
  \centering
  \includegraphics[width=0.91\linewidth]{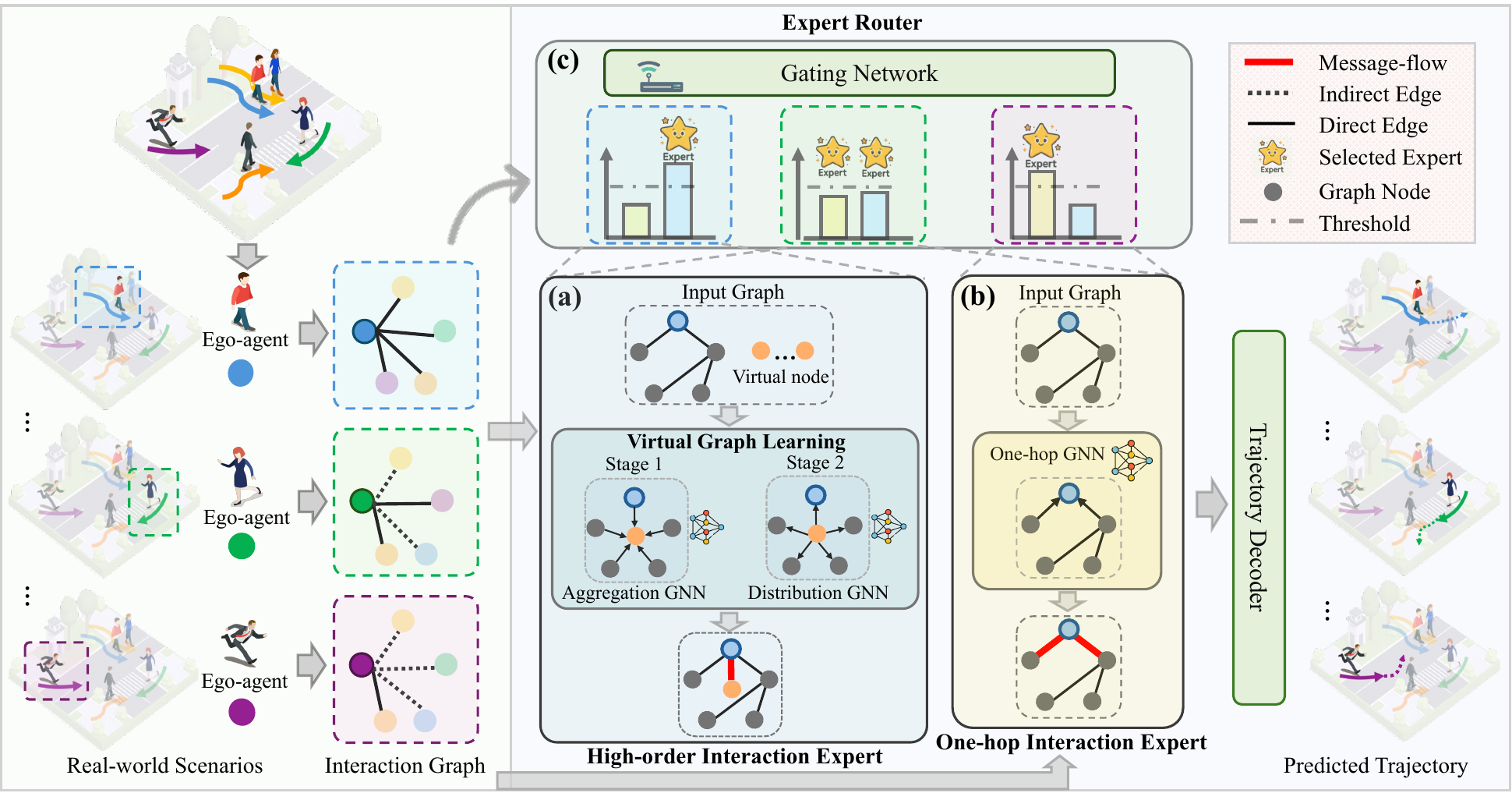}
  \caption{Overview of ViTE. Given pedestrian trajectories, we first construct interaction graphs. In (a), the high-order interaction expert captures indirect, long-range dependencies via Virtual Graph Learning, while (b) illustrates the one-hop expert modeling direct interactions. These expert outputs are then dynamically fused by a MoE-based Expert Router, as depicted in (c), enabling context-aware routing of graph information. Finally, an MLP-based decoder outputs future trajectories for each pedestrian.}
  \label{fig:overview}
\end{figure*}

\subsection{Problem Formulation}
Pedestrian trajectory prediction forecasts future positions from past movements. Mathematically, given a scene with $N$ pedestrians observed over $T_{obs}$ time steps, the trajectory of pedestrian $i$ is denoted as $\textit{X}_{i} = {(x_{t}^{i}, y_{t}^{i}) \mid t\in [-T_{obs}+1,\dots, 0]}$ for the observed past, and $\textit{Y}_{i} = { (x_{t}^{i}, y_{t}^{i}) \mid t\in [1,\dots, T_{pred}]}$ for the future ground-truth. Stacking all pedestrians yields the tensors $\mathbf{X} \in \mathbb{R}^{N \times T_{obs} \times 2}$ and $\mathbf{Y} \in \mathbb{R}^{N \times T_{pred} \times 2}$, representing the observed and future trajectories respectively, with each position in 2D coordinates. The goal is to minimize the error between the predicted trajectories $\hat{\mathbf{Y}}$ and ground-truth future trajectories $\mathbf{Y}$.

\subsection{Feature Initialization}
For each pedestrian $i$, we construct the input representation from two complementary feature types: absolute spatial coordinates $\mathbf{p}_{i} = (x, y)$ and temporal displacement vectors $\mathbf{r}_{i} = (r_x, r_y)$, where $\mathbf{r}_{i}^{(t)} = \mathbf{p}_{i}^{(t)} - \mathbf{p}_{i}^{(t-1)}$ captures the motion dynamics. The final input feature is obtained through concatenation: $\mathbf{X}_{i}^{\text{in}} = [\mathbf{p}_{i}; \mathbf{r}_{i}] \in \mathbb{R}^{T_{obs} \times 4}$. We represent pedestrian interactions using a spatial interaction graph $\mathcal{G} = (\mathcal{V}, \mathcal{E})$, where each node $n_i \in \mathcal{V}$ corresponds to pedestrian $i$, and each edge $e_{ij} \in \mathcal{E}$ models one-hop relationships. Node features are initialized as $\mathbf{n}_{i}^{(0)} = \mathcal{F}_{node}(\mathbf{X}_i^{\text{in}})$, where $\mathcal{F}_{node}(\cdot)$ denotes a Multi-Layer Perceptron (MLP). To capture social dependencies, we dynamically determine the graph connectivity via a $k$-nearest neighbor strategy based on feature similarity. For each connected pair $(i, j)$, we compute the initial embedding feature using a relational transformer (RT) layer \cite{diao2023relationaltrans,lee2024mart} equipped with sparse attention to ensure computational efficiency, denoted as $\mathbf{n}_{i,pair}^{0} = RT(\mathbf{n}_{i}^{(0)}, M)$, where $M$ is connection mask.

\subsection{High-order Interaction Modeling via Virtual Graph}

Capturing long-range dependencies is essential in trajectory prediction, as agents often influence one another beyond their immediate neighbors through high-order interactions. However, traditional GNNs based on first-order aggregation inherently suffer from \textit{under-reaching}—the inability to propagate information across distant nodes within shallow architectures \cite{qian2024probabilisticVN}. Although stacking GNN layers helps capture high-order interactions \cite{kim2024higher, chen2025pchgcn}, it significantly increases computational cost.

\begin{figure}[t]
  \centering
  \includegraphics[width=0.9\linewidth]{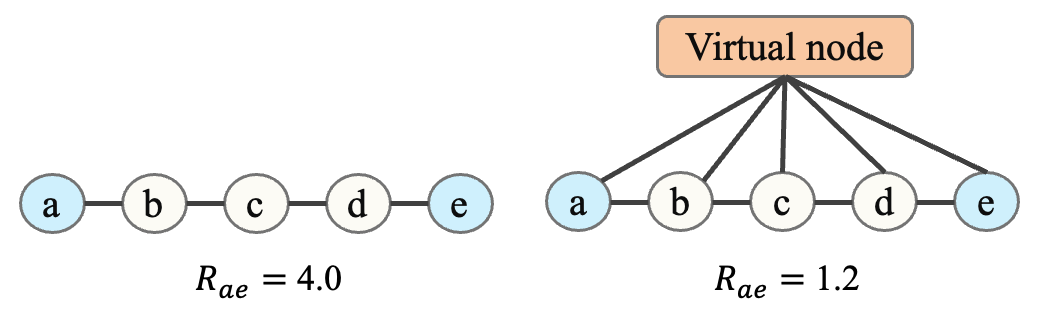}
  \caption{Comparison of effective resistance $(R_{ae})$ between a standard chain graph (left, $R_{ae} = 4.0$) and a virtual-node-enhanced graph structure (right, $R_{ae} = 1.2$). Lower effective resistance indicates more efficient message propagation and improved global connectivity.}
  \label{fig:effective_resistance}
\end{figure}

To quantify this limitation, we calculate the \textit{effective resistance} as a structural indicator of communication efficiency in graphs. Formally, the effective resistance between two nodes \(i\) and \(j\) is defined as:
\begin{equation}
R_{ij} = (e_i - e_j)^T \mathcal{L}^+ (e_i - e_j),
\end{equation}
where \(\mathcal{L}^+\) is the Moore–Penrose pseudoinverse of the graph Laplacian, and \(e_i, e_j\) are standard basis vectors. Lower \(R_{ij}\) indicates more efficient message propagation \cite{lu2024nodemixup}. For example, in the chain-structured graph shown in Figure~\ref{fig:effective_resistance} (left), node \(a\) requires four hops to reach node \(e\), resulting in a high resistance of \(R_{ae} = 4.0\), which hinders the efficiency of long-range communication. 

To address this issue, we propose the \textbf{Virtual Graph}, which introduces a small set of virtual nodes as mediators to facilitate high-order long-range communication. As shown in Figure~\ref{fig:effective_resistance} (right), adding a virtual node dramatically reduces the resistance between distant nodes (e.g., \(R_{ae} = 1.2\)), enabling more efficient global information flow.

Specifically, we construct a high-order interaction graph $\mathcal{G}_{\text{high}} = (\mathcal{N} \cup \mathcal{V}_{\text{virtual}}, \mathcal{E}_{\text{high}})$, where $\mathcal{N}$ denotes the set of real pedestrian nodes, and $\mathcal{V}_{\text{virtual}} = \{v_1, v_2, \dots, v_V\}$ is a set of virtual nodes acting as communication hubs. These virtual nodes are instantiated per training batch and initialized with diverse embeddings sampled from a learnable distribution:
\begin{equation}
\mathbf{v}_k^{(0)} \sim \mathcal{D}_{\text{vn}}, \quad k = 1,\dots,V,
\end{equation}
where $\mathcal{D}_{\text{vn}}$ is designed to promote representational diversity. To explicitly capture high-order dependencies, we propose a structured two-stage message passing scheme.

\textit{Stage 1 (Real-to-Virtual Message Aggregation).}
Each real node $n_i \in \mathcal{N}$ sends its node embedding to all virtual nodes, enabling them to aggregate diverse contexts into high-level representations. The embedding of each virtual node $v_k$ is updated as:
\begin{equation}
\mathbf{v}_{k}^{(1)} = \text{GNN}_{\text{r}\rightarrow\text{v}}\left(
\mathbf{v}_{k}^{(0)}, 
\text{AGG}_{\text{r}\rightarrow\text{v}}\left(\{\mathbf{n}_{i}^{(0)} \mid n_i \in \mathcal{N}\}\right)
\right),
\end{equation}
where $\text{AGG}_{\text{r}\rightarrow\text{v}}$ is a permutation-invariant aggregator (e.g., mean or attention), and $\text{GNN}_{\text{r}\rightarrow\text{v}}$ is a standard GNN module such as a Graph Attention Network \cite{petar2018GAT}.

\textit{Stage 2 (Virtual-to-Real Message Distribution).}
Each updated virtual node $v_k$ broadcasts its high-level contexts to all pedestrian nodes, enabling them to integrate indirect high-order interactions into their representations. The embedding of each pedestrian node $n_i$ is updated as:
\begin{equation}
\mathbf{n}_{i}^{(1)} = \text{GNN}_{\text{v}\rightarrow\text{r}}\left(
\mathbf{n}_{i}^{(0)}, 
\text{AGG}_{\text{v}\rightarrow\text{r}}\left(\{\mathbf{v}_{k}^{(1)} \mid v_k \in \mathcal{V}\}\right)
\right),
\end{equation}
where $\text{AGG}_{\text{v}\rightarrow\text{r}}$ is a permutation-invariant aggregator, and $\text{GNN}_{\text{v}\rightarrow\text{r}}$ is a standard GNN module for fusing virtual node information. The final output is then normalized and activated as: $\mathbf{n}_{i,\text{high}}^{(\text{out})} = \text{GELU}\left( \text{LayerNorm}\left(\mathbf{n}_{i}^{(1)}\right) \right)$.

In summary, our virtual graph interaction module integrates both localized and global context by combining one-hop and high-order message passing mechanisms. The one-hop graph captures fine-grained local interactions among neighboring agents, while the virtual-node-enhanced graph enables efficient modeling of high-order dependencies. Together, these complementary designs equip the model with a more expressive and scalable interaction representation.  

\begin{table*}[t]
\centering
\begingroup
\setlength{\tabcolsep}{1mm}
\small
\begin{tabular}{l|cccccccccc|c}
\toprule
\multicolumn{12}{c}{\textbf{ETH/UCY Dataset}} \\
\midrule
Subset 
& GroupNet & MemoNet & MID & NPSN & EqMotion & EigenTraj & LED & SingularTraj & MART & PCHGCN & Ours \\
\midrule
ETH    
& 0.46/0.73 & 0.40/0.61 & 0.39/0.66 & \underline{0.36}/0.59 & 0.40/0.61 & \underline{0.36}/0.53 & 0.39/0.58 & \textbf{0.35}/\textbf{0.42} & \textbf{0.35}/\underline{0.47} & 0.42/0.65 & \textbf{0.35}/0.49 \\

HOTEL  
& 0.15/0.25 & \textbf{0.11}/\textbf{0.17} & 0.13/0.22 & 0.16/0.25 & \underline{0.12}/\underline{0.18} & \underline{0.12}/0.19 & \textbf{0.11}/\textbf{0.17} & 0.13/0.19 & 0.14/0.22 & 0.17/0.28 & \textbf{0.11}/\textbf{0.17} \\

UNIV   
& 0.26/0.49 & 0.24/0.43 & \underline{0.22}/0.45 & 0.23/\underline{0.39} & 0.23/0.43 & 0.24/0.43 & 0.26/0.43 & 0.25/0.44 & 0.25/0.45 & \textbf{0.21}/\textbf{0.38} & 0.23/0.42 \\

ZARA1  
& 0.21/0.39 & \underline{0.18}/0.32 & \textbf{0.17}/0.30 & \underline{0.18}/0.32 & \underline{0.18}/0.32 & 0.19/0.33 & \underline{0.18}/\textbf{0.26} & 0.19/0.32 & \textbf{0.17}/\underline{0.29} & \textbf{0.17}/0.31 & \underline{0.18}/0.30 \\

ZARA2  
& 0.17/0.33 & \underline{0.14}/0.24 & \textbf{0.13}/0.27 & \underline{0.14}/0.25 & \textbf{0.13}/\underline{0.23} & \underline{0.14}/0.24 & \textbf{0.13}/\textbf{0.22} & 0.15/0.25 & \textbf{0.13}/\textbf{0.22} & \textbf{0.13}/\underline{0.23} & \textbf{0.13}/\textbf{0.22} \\
\midrule
AVG  
& 0.25/0.44 & \underline{0.21}/0.35 & \underline{0.21}/0.38 & \underline{0.21}/0.36 & \underline{0.21}/0.35 & \underline{0.21}/0.34 & \underline{0.21}/\underline{0.33} & \underline{0.21}/\textbf{0.32} & \underline{0.21}/\underline{0.33} & 0.22/0.37 & \textbf{0.20}/\textbf{0.32} \\
\bottomrule
\end{tabular}
\caption{Performance comparison on the ETH/UCY dataset. Metrics are $\min\mathrm{ADE}_{20}$/$\min\mathrm{FDE}_{20}$. \textbf{Bold} and \underline{underline} indicate the best and second-best results.}
\label{eth_results}
\endgroup
\end{table*}

\begin{table*}[t]
\centering
\begingroup
\setlength{\tabcolsep}{1.5mm}
\small
\begin{tabular}{l|ccccccccc|c}
\toprule
\multicolumn{11}{c}{\textbf{NBA Dataset}} \\
\midrule
Time
& STAR & GroupNet & MemoNet & MID & NPSN & DynGroupNet & LED & SingularTraj & MART & Ours \\
\midrule
1.0s & 0.43/0.66 & 0.26/0.34 & 0.38/0.56 & 0.28/0.37 & 0.35/0.58 & 0.19/0.28 & \underline{0.18}/0.27 & 0.28/0.44 & \underline{0.18}/\underline{0.26} & \textbf{0.17}/\textbf{0.25} \\
2.0s & 0.75/1.24 & 0.49/0.70 & 0.71/1.14 & 0.51/0.72 & 0.68/1.23 & 0.40/0.61 & \underline{0.37}/\underline{0.56} & 0.61/1.00 & \textbf{0.35}/\textbf{0.50} & \textbf{0.35}/\textbf{0.50} \\
3.0s & 1.03/1.51 & 0.73/1.02 & 1.00/1.57 & 0.71/0.98 & 1.01/1.76 & 0.65/0.90 & 0.58/0.84 & 0.96/1.47 & \underline{0.54}/\underline{0.71} & \textbf{0.53}/\textbf{0.70} \\
4.0s & 1.13/2.01 & 0.96/1.30 & 1.25/1.47 & 0.96/1.27 & 1.31/1.79 & 0.89/1.13 & 0.81/1.10 & 1.31/1.98 & \underline{0.73}/\textbf{0.90} & \textbf{0.72}/\underline{0.91} \\
\bottomrule
\end{tabular}
\caption{Performance comparison on the NBA dataset. Metrics are $\min\mathrm{ADE}_{20}$/$\min\mathrm{FDE}_{20}$. \textbf{Bold} and \underline{underline} indicate the best and second-best results.}
\label{nba_result}
\endgroup
\end{table*}

\subsection{One-hop Interaction Modeling}

To capture one-hop interactions among pedestrians, we define a first-order directed interaction graph $\mathcal{G}_{\text{one-hop}} = (\mathcal{N}, \mathcal{E}_{\text{one-hop}})$, where edges $\mathcal{E}_{\text{one-hop}}$ encode spatial relationships between neighboring agents based on their similarities \cite{lee2024mart}. This graph supports message passing operations in GNNs, enabling the modeling of one-hop dependencies among directly connected nodes \cite{ruochen2022multiclassSGCN,li2025bpsgcn}.

The expert leverages both node and edge features to update each node embedding. Specifically, each node $n_i$ aggregates information from its neighbors as follows:
\begin{equation}
\mathbf{m}_{ij} = \phi_m\left(\mathbf{n}_i^{(0)}, \mathbf{n}_j^{(0)}, \mathbf{e}_{ij}^{(0)}\right),
\end{equation}
\begin{equation}
\mathbf{m}_i = \text{AGG}_{\text{one-hop}}\left(\{\mathbf{m}_{ij} \mid j \in \mathcal{N}_{\text{one-hop}}(i)\}\right),
\end{equation}
\begin{equation}
\mathbf{n}_{i,\text{one-hop}}^{(\text{out})} = \phi_u\left(\mathbf{n}_i^{(0)}, \mathbf{m}_i\right),
\end{equation}
where $\phi_m(\cdot)$ denotes the message function that integrates source node, target node, and edge features; $\text{AGG}_{\text{one-hop}}(\cdot)$ is a permutation-invariant aggregation function; and $\phi_u(\cdot)$ is the update function, typically an MLP layer. This design allows the model to leverage both one-hop structural information and relation features for direct interaction modeling.

\subsection{Expert Router}

While one-hop and high-order graph structures capture complementary interaction patterns, existing methods typically adopt a fixed graph design, limiting adaptability to varying scene complexities. Such rigidity can result in under-reaching in complex scenarios or redundant computation in simpler ones. To address this limitation, we propose an adaptive \textbf{Expert Router} based on the MoE paradigm, which learns a context-aware routing policy to dynamically allocate interaction modeling capacity. Specifically, we treat $\mathcal{G}_{\text{one-hop}}$ and $\mathcal{G}_{\text{high}}$ as two specialized experts responsible for modeling direct and high-order interactions, respectively. By allowing each node to selectively attend to the most relevant expert based on its feature representations.

For each node feature $\mathbf{n}_i^{(0)}$, the gating network computes a soft routing distribution \cite{hu2025timefilter} over the two experts:
\begin{gather}
\mathbf{g}_i = \text{Softmax}(\phi(\mathbf{n}_i^{(0)})), \label{eq:gating}\\
\phi(\mathbf{n}_i^{(0)}) = \mathbf{W}_g \mathbf{n}_i^{(0)} + \boldsymbol{\epsilon} \cdot \text{Softplus}(\mathbf{W}_n \mathbf{n}_i^{(0)}), \label{eq:noise}
\end{gather}
where $\phi(\mathbf{n}_i^{(0)}) \in \mathbb{R}^2$ are the logits over two experts, and $\boldsymbol{\epsilon} \sim \mathcal{N}(\mathbf{0}, \mathbf{I})$ adds Gaussian noise for regularization. $\mathbf{W}_g$ and $\mathbf{W}_n$ are learnable projection matrices.

To enable adaptive complexity control and improve efficiency, we adopt a threshold-based Top-P mechanism that dynamically selects active experts \cite{huang2024harderMOE,hu2025timefilter}. Unlike traditional MoE methods \cite{wang2023graphMOE} that always fuse all experts, our approach activates experts based on confidence, enabling lightweight local processing for simple cases and triggering global reasoning in complex interaction scenarios.

Let $\tilde{\mathbf{g}}_i = \text{Sort}(\mathbf{g}_i, \text{descending})$ be the sorted expert probabilities. The active expert set $\mathcal{S}_i$ is defined as the minimal set whose cumulative probability exceeds a threshold $p$:
\begin{equation}
\mathcal{S}_i = \left\{k : \sum_{j=1}^{k} \tilde{g}_{i,j} \leq p \right\} \cup \left\{ \arg\min_{k} \left\{\sum_{j=1}^{k} \tilde{g}_{i,j} > p \right\} \right\},
\label{eq:top_p}
\end{equation}
where $p \in (0, 1)$ controls the sparsity level (e.g., $p = 0.7$). The selected expert weights are then renormalized as:
\begin{equation}
\hat{g}_{i,k} = \begin{cases}
\frac{g_{i,k}}{\sum_{j \in \mathcal{S}_i} g_{i,j}} & \text{if } k \in \mathcal{S}_i, \\
0 & \text{otherwise}.
\end{cases}
\label{eq:renorm}
\end{equation}

Given the selected expert set $\mathcal{S}_i$ and renormalized weights $\hat{g}_{i,k}$ from Eq.~\ref{eq:renorm}, the final node representation is computed as a weighted sum:
\begin{equation}
\mathbf{n}_{i,router}^{(\text{out})} = \hat{g}_{i,1} \cdot \mathbf{n}_{i,\text{one-hop}}^{(\text{out})} + \hat{g}_{i,2} \cdot \mathbf{n}_{i,\text{high}}^{(\text{out})}.
\label{eq:final_fusion}
\end{equation}

By leveraging soft gating and selective expert activation, the proposed \textbf{Expert Router} enables context-aware routing of one-hop and high-order interactions, allowing the model to dynamically adjust its reasoning complexity based on pedestrian behavior patterns.

To prevent routing collapse and promote expert diversity, we introduce an importance-based auxiliary loss \cite{hu2025timefilter}:
\begin{equation}
\mathcal{L}_{\text{imp}} = \frac{1}{N} \sum_{i=1}^{N} \frac{\text{Std}(\mathbf{g}_i)}{\text{Mean}(\mathbf{g}_i) + \epsilon}, \label{eq:importance}
\end{equation}
which encourages balanced expert utilization across nodes and improves the robustness of the routing mechanism.

\subsection{Trajectory Decoding}

For trajectory prediction, we concatenate three feature types $[\mathbf{n}_{i}^{(0)}; \mathbf{n}_{i,pair}^{(0)}; \mathbf{n}_{i,router}^{(\text{out})}]$ and feed them into $K$ parallel MLP-based prediction heads to generate diverse trajectories following \cite{xu2023eqmotion,lee2024mart}. Given $K$ predicted trajectories $\hat{\mathbf{Y}}_{i}^{(k)}$ from $K$ prediction heads for pedestrian $i$, we compute the prediction loss by selecting the best trajectory in terms of minimum $\ell_2$ distance to the ground truth:
\begin{equation}
\mathcal{L}_{\text{pred}} = \frac{1}{N T_{pred}} \sum_{i=1}^N \sum_{t=1}^{T_{pred}} \min_{k} || \mathbf{p}_{i}^{(t)} - \hat{\mathbf{p}}_{i}^{(t,k)} ||_{2},
\end{equation}
where $\mathbf{p}_{i}^{(t)} = (x_t^i, y_t^i)$ is the ground-truth position of pedestrian $i$ at time $t$, and $\hat{\mathbf{p}}_{i}^{(t,k)}$ is the corresponding prediction from the $k$-th head. The total loss incorporates MoE regularization: $\mathcal{L} = \mathcal{L}_{\text{pred}} + \lambda \mathcal{L}_{\text{imp}}$.

\begin{table*}[t]
\centering
\begingroup
\setlength{\tabcolsep}{1mm}
\small
\begin{tabular}{l|ccccccccc|c}
\toprule
\multicolumn{11}{c}{\textbf{SDD Dataset}} \\
\midrule
Time 
& PECNet & GroupNet & MemoNet & MID & NPSN & DynGroupNet & EigenTraj & LED & MART & Ours \\
\midrule
4.8s 
& 9.96/15.88 
& 9.31/16.11 
& 8.56/12.66 
& 9.73/15.32 
& 8.56/11.85 
& 8.42/13.58 
& 8.05/13.25 
& 8.48/\textbf{11.66} 
& \underline{7.43}/\underline{11.82} 
& \textbf{7.42}/11.90 \\
\bottomrule
\end{tabular}
\caption{Performance comparison on the SDD dataset. Metrics are $\min\mathrm{ADE}_{20}$/$\min\mathrm{FDE}_{20}$. \textbf{Bold} and \underline{underline} indicate the best and second-best results.}
\label{sdd_result}
\endgroup
\end{table*}

\section{Experiments}

\subsection{Experimental Setup}

\subsubsection{Datasets.}
We evaluate our model on three widely-used trajectory prediction benchmarks: ETH/UCY \cite{pellegrini2009ETH,Lerner2007UCY}, Stanford Drone Dataset (SDD) \cite{Robicquet2016SDD}, and NBA SportVU \cite{Leapfrog}. The ETH/UCY dataset comprises five subsets (ETH, HOTEL, UNIV, ZARA1, and ZARA2), capturing pedestrian movements across diverse social scenarios. SDD is a large-scale dataset collected from a university campus. For both ETH/UCY and SDD, we follow the standard setting in \cite{lee2024mart,xu2022remembermemo}, using 3.2 seconds (8 frames) of observed trajectories to predict the next 4.8 seconds (12 frames). For ETH/UCY, we adopt a leave-one-out training protocol, training on four subsets and testing on the remaining one. The NBA SportVU dataset provides trajectories of 10 players and the ball in real basketball games. We follow \cite{Xu2022GroupNetMH,Leapfrog,lee2024mart} that use 2.0 seconds (10 frames) of past motion to predict the next 4.0 seconds (20 frames).

\subsubsection{Evaluation Metrics.}
To evaluate the performance, we adopt two metrics: $\min\mathrm{ADE}_k$ and $\min\mathrm{FDE}_k$, following the evaluation protocol in \cite{gupta2018socialgan,lee2024mart}. The Average Displacement Error (ADE) measures the mean Euclidean distance between the predicted and ground-truth trajectories over all time steps, while the Final Displacement Error (FDE) focuses on the distance between the predicted and actual final positions. Given $k$ sampled predictions for each agent, we report the minimum $\mathrm{ADE}$ and $\mathrm{FDE}$ among them, denoted as $\min\mathrm{ADE}_k$ and $\min\mathrm{FDE}_k$.

\subsection{Comparison with State-of-the-Art Methods}

The results on the ETH/UCY dataset are presented in Table \ref{eth_results}, our method achieves the best overall performance, with the lowest average ADE/FDE of 0.20/0.32 across all five subsets. Compared to PCHGCN, a high-order graph-based method, our approach reduces ADE by 9.1\% and FDE by 13.5\%. Moreover, our method consistently ranks among the top performers on each individual subset, demonstrating strong generalization across diverse crowd scenarios. For NBA datasets shown in Table \ref{nba_result}, our model ranks first at 1.0s, 2.0s, and 3.0s, and achieves the lowest ADE and second-best FDE at 4.0s, highlighting its robustness in modeling long-term multi-agent interactions in dynamic sports environments. Table \ref{sdd_result} presents the results on the SDD dataset. Our method achieves the best ADE and a competitive FDE compared to existing approaches, confirming its effectiveness in forecasting pedestrian trajectories in real-world scenes.

\subsection{Ablation Study and Analysis}
\subsubsection{Effect of Expert.}
We conduct ablation studies to assess the impact of different configurations of Expert Router (Table~\ref{expert_ablation}). Removing both experts results in a clear performance drop, confirming the necessity of incorporating expert mechanisms into the model. When comparing single-expert settings, the one-hop expert yields better results than the high-order counterpart, indicating that certain interaction structures are more informative for prediction. Most notably, our adaptive MoE outperforms both individual experts and the fixed mixed configuration, demonstrating that dynamic, input-dependent expert selection offers clear advantages over static alternatives. These findings validate the effectiveness of context-aware expert routing in modeling diverse trajectory patterns.

\begin{table}[t]
\centering
\begingroup
\setlength{\tabcolsep}{4pt}
\small
\begin{tabular}{l|c c}
\toprule
\textbf{Expert Configuration} & \multicolumn{2}{c}{\textbf{ETH/UCY Dataset}} \\
\cmidrule(lr){2-3}
& $\min\mathrm{ADE}_{20}$ & $\min\mathrm{FDE}_{20}$ \\
\midrule
One-hop Expert Only    & \underline{0.22} & \underline{0.34} \\
High-order Expert Only & 0.23 & 0.36 \\
Mixed Expert      & 0.25 & 0.36 \\
No Expert         & 0.25 & 0.38 \\
\hline
Ours              & \textbf{0.20} & \textbf{0.32} \\
\bottomrule
\end{tabular}
\caption{Ablation study of expert selection evaluated by $\min\mathrm{ADE}_{20}$/$\min\mathrm{FDE}_{20}$ on ETH/UCY dataset. \textbf{Bold} and \underline{underline} indicate the best and second-best results.}
\label{expert_ablation}

\endgroup
\end{table}

\begin{table}[t]
\centering
\begingroup
\setlength{\tabcolsep}{3mm}
\small
\begin{tabular}{l|cc}
\toprule
Method       & \#Param. & MACs \\
\midrule
PECNet      & 2.1M   & 259.2M \\
STAR        & \textbf{1.0}M   & 12.0G \\
MemoNet     & 10.7M  & 6.0G \\
GroupNet    & 2.2M   & 411.5M \\
MID         & 9.0M   & 40.3G \\
EqMotion    & 3.0M   & 147.1M \\
LED         & 10.9M  & 15.0G \\
MART        & \underline{1.5}M   & \underline{43.3}M \\
\hline
Ours   & \textbf{1.0}M & \textbf{23.1}M \\
\bottomrule
\end{tabular}
\caption{Model complexity comparison. \textbf{Bold} and \underline{underline} indicate the best and second-best results.}
\label{complexity_analysis}
\endgroup
\end{table}

\begin{figure*}[t]
\centering
\includegraphics[width=0.78\linewidth]{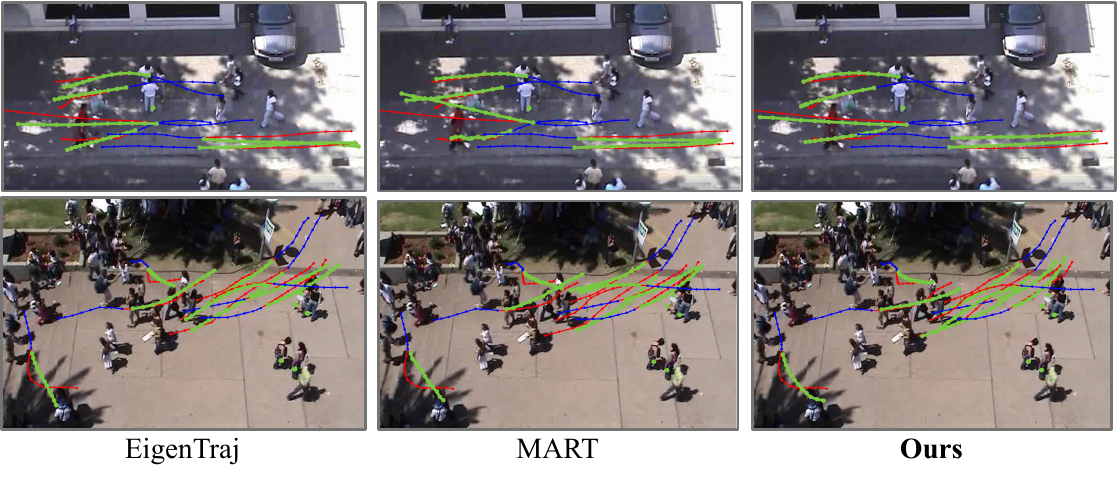} 
\caption{Qualitative results on ETH/UCY datasets. Historical trajectories are in blue, ground-truth trajectories are in red, and predicted trajectories are in green.}
\label{fig:ethucy}
\end{figure*}

\begin{figure}[t]
\centering
\includegraphics[width=\columnwidth]{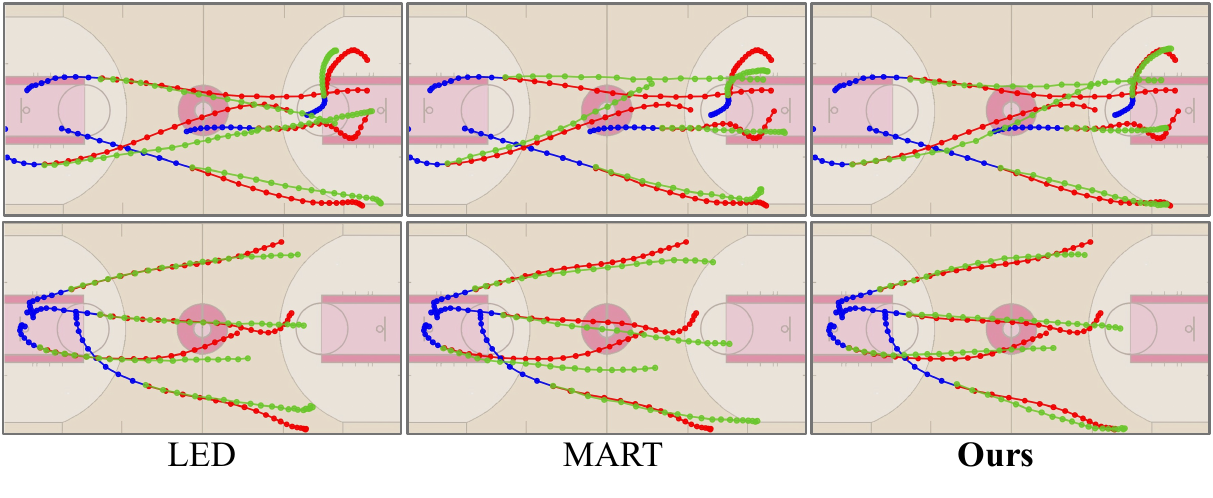} 
\caption{Qualitative results on NBA datasets. Historical trajectories are in blue, ground-truth trajectories are in red, and
predicted trajectories are in green.}
\label{fig:nba}
\end{figure}

\begin{figure}[t]
\centering
\includegraphics[width=\columnwidth]{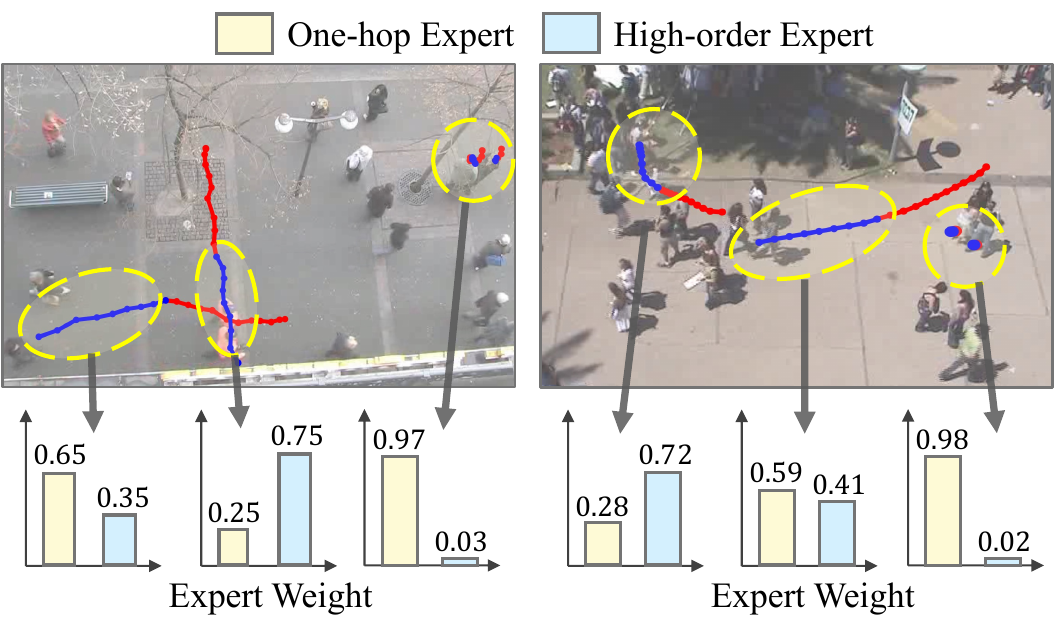} 
\caption{Visualization of learned expert weights across different scenarios. The model dynamically adjusts one-hop and high-order expert contributions by interaction complexity.}
\label{fig:expert_weight}
\end{figure}

\subsubsection{Effect of Virtual Nodes.} To assess the effectiveness of the Virtual Graph, we compare our model with conventional multi-layer GCNs. As shown in Table \ref{gcn_ablation}, our approach achieves the best performance with the fewest parameters (1.00×), outperforming both 2-layer and 4-layer GCNs. Notably, simply increasing GCN depth does not yield better results—the 4-layer model even slightly underperforms the 2-layer variant, despite having more parameters. This suggests that stacking layers is neither efficient nor sufficient for capturing high-order interactions. In contrast, our use of virtual nodes offers a more effective and parameter-efficient solution for modeling such interactions.

\begin{table}[t]
\centering
\begingroup
\setlength{\tabcolsep}{4pt}
\small
\begin{tabular}{l|ccc}
\toprule
\textbf{Variant} & \multicolumn{3}{c}{\textbf{ETH/UCY Dataset}} \\
\cmidrule(lr){2-4}
& $\min\mathrm{ADE}_{20}$ & $\min\mathrm{FDE}_{20}$ & Param. Ratio \\
\midrule
GCN (2-layer) & \underline{0.22} & \underline{0.36} & 0.97x \\
GCN (4-layer)   & 0.23 & \underline{0.36} & 1.11x \\
\hline
Ours        & \textbf{0.20} & \textbf{0.32} & 1.00x \\
\bottomrule
\end{tabular}
\caption{Ablation study on virtual graph effectiveness evaluated by $\min\mathrm{ADE}_{20}$/$\min\mathrm{FDE}_{20}$ on ETH/UCY dataset. \textbf{Bold} and \underline{underline} indicate the best and second-best results.}
\label{gcn_ablation}
\endgroup
\end{table}

\subsubsection{Efficiency Comparisons.} Table \ref{complexity_analysis} compares model complexity in terms of parameters and Multiply–Accumulate Operations (MACs). Following the evaluation protocol of \cite{lee2024mart}, we compute MACs using scenes with 10 agents from the ETH/UCY dataset. Our method achieves the lowest computational complexity (23.1M MACs) and the smallest parameter count (1.0M) among all baselines. These results highlight the efficiency of our model and its suitability for real-world deployment.

\subsection{Qualitative Results}
\subsubsection{Visualization of Predicted Trajectory.} Figure \ref{fig:ethucy} shows qualitative comparisons on the ETH/UCY dataset among EigenTraj, MART, and our model. Our predictions (green) align more closely with ground-truth trajectories (red), capturing pedestrians' subtle movements and social interactions more accurately. Figure \ref{fig:nba} presents trajectory predictions on the NBA dataset comparing LED, MART, and our approach. Our method consistently generates smoother and more precise trajectories, effectively capturing complex dynamics and long-term interactions among multiple agents. These visual results further confirm our model's superior predictive capability across diverse scenarios.

\subsubsection{Expert Weight Analysis.} Figure \ref{fig:expert_weight} illustrates the learned expert weights for different trajectory scenarios. The results indicate that the one-hop expert generally receives higher weights in simpler interactions. Conversely, the high-order expert dominates in more complex, globally interactive scenarios. This adaptive expert weighting confirms the effectiveness of our model in dynamically balancing the usage of one-hop expert and high-order expert.

\section{Conclusion}
We propose ViTE, a novel trajectory prediction framework combining Virtual Graph for high-order interactions and Expert Router for adaptive expert routing. This design balances multi-scale reasoning and computational efficiency. Experiments on ETH/UCY, NBA, and SDD confirm state-of-the-art performance. Future work will incorporate contextual image information to further enhance scene understanding. Integrating visual semantics such as obstacles, road structure, or group behavior cues could provide stronger priors for trajectory prediction. We also aim to explore more flexible expert architectures that dynamically adjust their granularity or number based on scene complexity.

\section{Acknowledgment}
{This project is supported in part by the EPSRC NortHFutures project (ref: EP/X031012/1).}

\bibliography{Bibliography-File}



\section*{Supplementary material}

# Supplementary Material for
# ViTE: Virtual Graph Trajectory Expert Router for Pedestrian Trajectory Prediction

Ruochen Li[1], Zhanxing Zhu[2], Tanqiu Qiao[1], Hubert P. H. Shum[1]*

[1]Department of Computer Science, Durham University, UK
[2]School of Electrical and Computer Science (ECS), University of Southampton, UK
{ruochen.li, hubert.shum}@durham.ac.uk

This supplementary document offers additional technical and experimental details that support the findings reported in the main paper. We begin with an overview of the baseline methods used for comparison. We then describe our implementation details. Finally, we present supplementary experiments and qualitative results to further illustrate the effectiveness of our proposed framework.

## Details of Baselines

- **STAR** (Yu et al. 2020): This model introduces a spatial-temporal graph transformer framework focusing on trajectory prediction.
- **PECNet** (Mangalam et al. 2020): This model proposes a predicted endpoint-conditioned network that captures stochastic goals and introduces a novel non-local social pooling layer for human trajectory prediction.
- **GroupNet** (Xu et al. 2022a): This method proposes a multiscale hypergraph neural network for multi-agent trajectory prediction.
- **MemoNet** (Xu et al. 2022b): MemoNet is an instance-based trajectory prediction framework that mimics human memory by retrieving similar past scenarios from a memory bank to improve prediction performance.
- **MID** (Gu et al. 2022): This method introduces a parameterized Markov chain and a transformer-based diffusion model for trajectory prediction
- **NPSN** (Bae, Park, and Jeon 2022): This method tackles trajectory variability using Quasi-Monte Carlo sampling and introduces the Non-Probability Sampling Network (NPSN) for improved prediction.
- **DynGroupNet** (Xu et al. 2024): This paper introduces a dynamic-group-aware model improves interaction modeling for trajectory prediction.
- **EigenTraj** (Bae, Oh, and Jeon 2023): The paper presents EigenTrajectory (EigenTraj), a method for enhancing pedestrian trajectory prediction by constructing a compact ET space to represent movement patterns.
- **EqMotion** (Xu et al. 2023): EqMotion predicts trajectories with Euclidean equivariance and models interactions in a transformation-invariant manner.
- **LED** (Mao et al. 2023): LED is a diffusion-based trajectory predictor that generates diverse outputs efficiently by using a leapfrog initializer to skip denoising steps.
- **SingularTraj** (Bae, Park, and Jeon 2024): The paper presents SingularTrajectory (SingularTraj), a diffusion-based framework for pedestrian trajectory prediction.
- **MART** (Lee et al. 2024): This paper introduces multi-scale relational transformer for trajectory prediction.
- **PCHGCN** (Chen, Sang, and Zhao 2025): This paper introduces a high-order spatial-temporal framework for pedestrian trajectory prediction.

## Implementation Details

As shown in Table 1, we summarize the key model components and training hyperparameters used in our ViTE framework. In the one-hop interaction expert, we adopt a two-layer Graph Convolutional Network (GCN) (Kipf and Welling 2017) to model local interactions. For the virtual graph module, the first stage employs a single-layer Graph Attention Network (GAT) (Veličković et al. 2018) for feature aggregation, while the second stage uses a single-layer GCN for feature distribution. We use the MultiStepLR scheduler with a decay factor of 0.5 to adjust the learning rate during training. For ETH/UCY datasets, we apply data augmentation following (Mohamed et al. 2022). The model is trained on a single NVIDIA TITAN Xp GPU for each dataset with Adam optimizer.

## Supplementary Analysis

### Virtual Node Mechanism Analysis

Figure 1 (left) shows the effect of virtual node count on prediction accuracy. We observe that using only one virtual node leads to suboptimal performance, likely due to limited capacity for modeling diverse global interactions. Performance improves and peaks when using three virtual nodes, indicating this configuration offers a good balance between expressiveness and efficiency. Adding more virtual nodes (e.g., four or five) results in slight degradation, suggesting that excessive

---
*Corresponding author

| Parameter | ETH | HOTEL | UNIV | ZARA1 | ZARA2 | NBA | SDD |
|---|---|---|---|---|---|---|---|
| Learning Rate | 0.001 | 0.0018 | 0.001 | 0.0012 | 0.0012 | 0.0005 | 0.001 |
| Training Epochs | 300 | 300 | 300 | 300 | 300 | 100 | 300 |
| Top-P Threshold | 0.5 | 0.6 | 0.5 | 0.6 | 0.5 | 0.5 | 0.5 |
| # Expert Router Layers | 3 | 3 | 3 | 3 | 3 | 2 | 1 |
| # Virtual Nodes | 3 | 3 | 3 | 3 | 3 | 2 | 1 |

Table 1: Hyperparameter settings for each dataset.

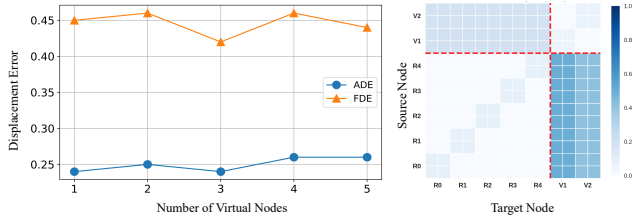

Figure 1: Analysis of virtual node mechanisms. Prediction accuracy (ADE/FDE) with varying virtual node counts on UNIV subset (**Left**). Attention map for message-passing between real nodes (R0–R4) and virtual nodes (V1–V2) (**Right**).

nodes may introduce redundant or noisy interactions that hinder learning.

To better understand how virtual nodes function, we visualize the attention weights between real nodes (R0–R4) and virtual nodes (V1–V2) in Figure 1 (right). The attention map reveals that both virtual nodes maintain strong connections with all real agents, confirming their role as global aggregators. Notably, V1 and V2 exhibit different attention patterns, indicating complementary specialization. The asymmetric structure of the map—where information flows from real nodes to virtual nodes and then back—also reflects our two-stage design, enabling virtual nodes to collect and redistribute high-level context across the graph.

## More Qualitative Results

Figure 2 presents qualitative results of our framework across the five ETH/UCY subsets, arranged from left to right in order of increasing scene complexity. These scenarios exhibit diverse and stochastic pedestrian movement behaviors, including nonlinear movements, abrupt direction changes, and densely entangled trajectories. Despite such challenges, our model consistently produces accurate and socially compliant predictions that closely follow ground-truth trajectories.

In relatively simple environments such as **ETH** and **HOTEL**, the predicted trajectories align smoothly with historical trajectories. For mid-level complexity scenes like **ZARA1**, where subtle group interactions and moderate scene dynamics are present, our model effectively maintains prediction accuracy while preserving socially aware movement patterns. In the more complex subsets **ZARA2** and **UNIV**, which contain frequent agent crossings and rapid directional changes, our approach remains robust, accurately anticipating individual behaviors and inter-agent spacing. These results highlight the strong generalization ability of our expert-driven interaction modeling across both structured and highly dynamic pedestrian scenarios.

## Failure Case Discussion

While our model demonstrates strong overall performance across diverse scenes, we observe certain failure cases in highly interactive or converging pedestrian scenarios. As shown in Figure 3, when multiple agents approach a shared goal region or intersecting paths, the model may predict future trajectories that lead to potential collisions. This issue often occurs when agents have ambiguous or overlapping motion patterns during the observation period. Without explicit collision avoidance or intent modeling, the model may overfit to historical movements and fail to anticipate social negotiation. While our interaction modeling performs well in general, future work could explore collision-aware objectives or intent-aware prediction to improve physical plausibility and social compliance.

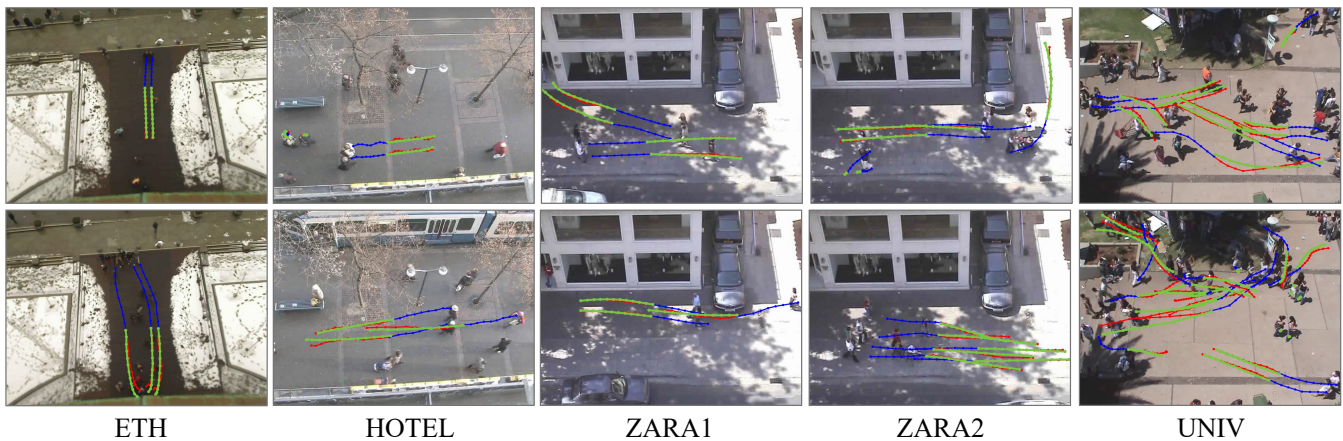

Figure 2: Qualitative results on ETH/UCY datasets. Historical trajectories are in blue, ground-truth trajectories are in red, and predicted trajectories are in green.

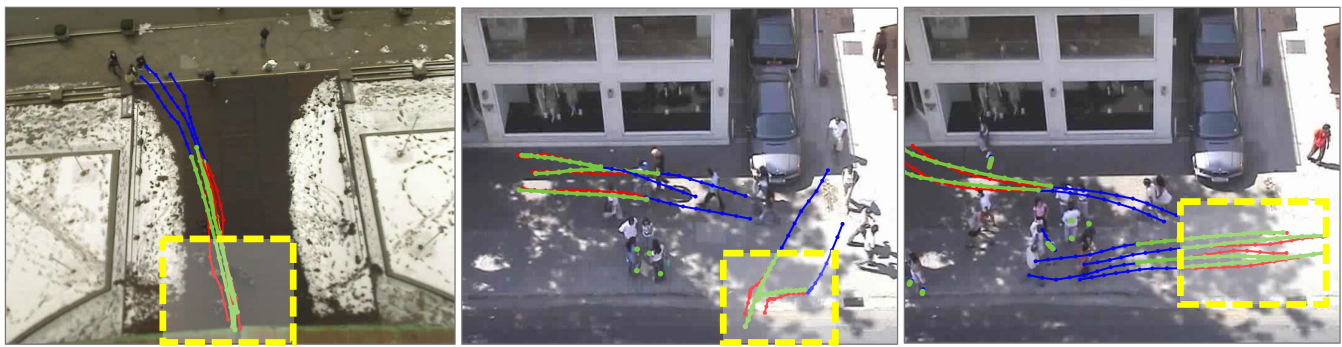

Figure 3: Illustration of a failure case. Historical trajectories are in blue, ground-truth trajectories are in red, and predicted trajectories are in green. Regions with potential trajectory overlap are highlighted by yellow boxes.



\end{document}